\documentclass[runningheads]{llncs}
\usepackage{graphicx}
\usepackage{times}
\usepackage{epsfig}
\usepackage{graphicx}
\usepackage{amsmath}
\usepackage{amssymb}
\usepackage{multirow}
\usepackage{xspace}
\usepackage{xcolor}
\usepackage{url}
\usepackage{hyperref}

\newcommand{\modelname}[1]{\textsc{#1}}
\newcommand{\etal}{\textit{et al.}}

\usepackage[misc]{ifsym}
\begin{document}
\title{Automatic Recognition of Student Engagement using Deep Learning and Facial Expression}
\titlerunning{Automatic Recognition of Student Engagement using Deep Learning}
%

\author{Omid Mohamad Nezami\inst{1,2} (\Letter)
\and
Mark Dras\inst{1} \and Len Hamey\inst{1} \and
Deborah Richards\inst{1} Stephen Wan\inst{2} \and C\'ecile Paris\inst{2}}

\authorrunning{Nezami et al.}
%
\institute{Macquarie University,
Sydney, NSW, Australia\\
\email{omid.mohamad-nezami@hdr.mq.edu.au}\\
\email{\{mark.dras,len.hamey,deborah.richards\}@mq.edu.au}\\
\and
CSIRO's Data61, Sydney, NSW, Australia\\
\email{\{stephen.wan,cecile.paris\}@data61.csiro.au}}

%
\maketitle              
\begin{abstract}
Engagement is a key indicator of the quality of learning experience, and one that plays a major role in developing intelligent educational interfaces. Any such interface requires the ability to recognise the level of engagement in order to respond appropriately; however, there is very little existing data to learn from, and new data is expensive and difficult to acquire. This paper presents a deep learning model to improve engagement recognition from images that overcomes the data sparsity challenge by pre-training on readily available basic facial expression data, before training on specialised engagement data. In the first of two steps, a facial expression recognition model is trained to provide a rich face representation using deep learning. In the second step, we use the model's weights to initialize our deep learning based model to recognize engagement; we term this the engagement model. We train the model on our new engagement recognition dataset with 4627 engaged and disengaged samples. We find that the engagement model outperforms effective deep learning architectures that we apply for the first time to engagement recognition, as well as approaches using histogram of oriented gradients and support vector machines.

\keywords{Engagement \and Deep Learning \and Facial Expression.}
\end{abstract}
\section{Introduction}

Engagement is a significant aspect of human-technology interactions and is defined differently for a variety of applications such as search engines, online gaming platforms, and mobile health applications \cite{o2016theoretical}. According to Monkaresi~\etal~\cite{monkaresi2017automated}, most definitions describe engagement as attentional and emotional involvement in a task.

\begin{figure}[t]
    \begin{center}
    \includegraphics[width=0.5\linewidth]{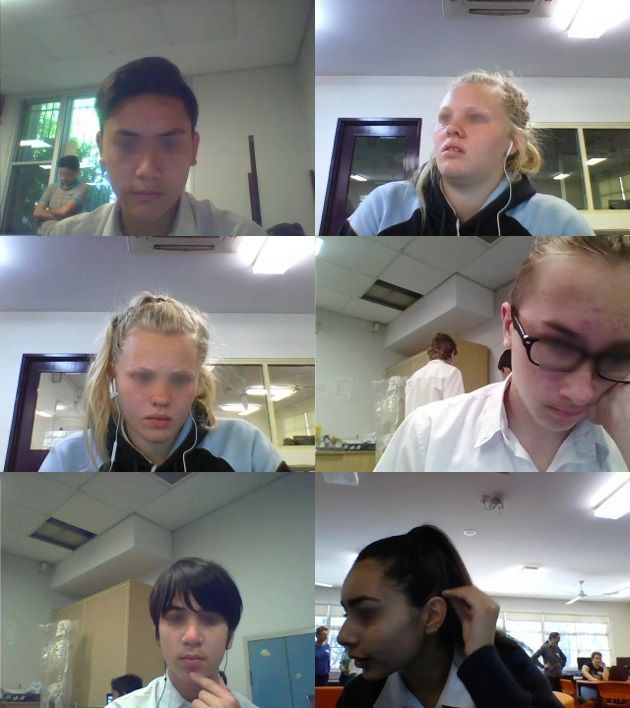}
    \end{center}
    \caption{Engaged (left) and disengaged (right) samples collected in our studies. We blurred the children's eyes for ethical issues, even though we have their parents consent at the time.}
    \label{fig:samples}
\end{figure}

This paper deals with engagement during learning via technology. Investigating engagement is vital for designing intelligent educational interfaces in different learning settings including educational games \cite{jacobson2016computational}, massively open online courses (MOOCs) \cite{kamath2016crowdsourced}, and intelligent tutoring systems (ITSs) \cite{alyuz2016semi}. For instance, if students feel frustrated and become disengaged (see disengaged samples in Fig.~\ref{fig:samples}), the system should intervene in order to bring them back to the learning process. However, if students are engaged and enjoying their tasks (see engaged samples in Fig.~\ref{fig:samples}), they should not be interrupted even if they are making some mistakes \cite{kapoor2001towards}. In order for the learning system to adapt the learning setting and provide proper responses to students, we first need to automatically measure engagement. This can be done by, for example, using context performance \cite{alyuz2016semi}, facial expression \cite{whitehill2014faces} and heart rate \cite{monkaresi2017automated} data. Recently, engagement recognition using facial expression data has attracted special attention because of widespread availability of cameras~\cite{monkaresi2017automated}.

This paper aims at quantifying and characterizing engagement using facial expressions extracted from images. In this domain, engagement detection models usually use typical facial features which are designed for general purposes, such as Gabor features \cite{whitehill2014faces}, histogram of oriented gradients \cite{kamath2016crowdsourced} and facial action units \cite{bosch2015automatic}. To the best of the authors' knowledge, there is no work in the literature investigating the design of specific and high-level features for engagement. Therefore, providing a rich engagement representation model to distinguish engaged and disengaged samples remains an open problem (Challenge 1). Training such a rich model requires a large amount of data which means extensive effort, time, and expense would be required for collecting and annotating data due to the complexities \cite{bosch2016detecting} and ambiguities \cite{o2016theoretical} of the engagement concept (Challenge 2).

To address the aforementioned challenges, we design a deep learning model which includes two essential steps: basic facial expression recognition, and engagement recognition. In the first step, a convolutional neural network (CNN) is trained on the dataset of the Facial Expression Recognition Challenge 2013 (FER-2013) to provide a rich facial representation model, achieving state-of-the-art performance. In the next step, the model is applied to initialize our engagement recognition model, designed using a separate CNN, learned on our newly collected dataset in the engagement recognition domain. As a solution to Challenge 1, we train a deep learning-based model that provides our representation model specifically for engagement recognition. As a solution to Challenge 2, we use the FER-2013 dataset, which is around eight times larger than our collected dataset, as external data to pre-train our engagement recognition model and compensate for the shortage of engagement data~\footnote{Our code and trained models are publicly available from \url{https://github.com/omidmnezami/Engagement-Recognition}}. The contributions of this work are threefold:

\begin{itemize}
\item
To the authors' knowledge, the work in this paper is the first time a rich face representation model has been used to capture basic facial expressions and initialize an engagement recognition model, resulting in positive outcomes. This shows the effectiveness of applying basic facial expression data in order to recognize engagement.
\item
We have collected a new dataset we call the Engagement Recognition (ER) dataset to facilitate research on engagement recognition from images. To handle the complexity and ambiguity of engagement concept, our data is annotated in two steps, separating the behavioral and emotional dimensions of engagement. The final engagement label in the ER dataset is the combination of the two dimensions.
\item
To the authors' knowledge, this is the first study which models engagement using deep learning techniques. The proposed model outperforms a comprehensive range of baseline approaches on the ER dataset.
\end{itemize}


\section{Related Work}
\label{sec:relwork}
\subsection{Facial Expression Recognition}
As a form of non-verbal communication, facial expressions convey attitudes, affects, and intentions of people. They are the result of movements of muscles and facial features \cite{fasel2003automatic}. Study of facial expressions was started more than a century ago by Charles Darwin \cite{ekman2006darwin}, leading to a large body of work in recognizing basic facial expressions \cite{fasel2003automatic,sariyanidi2015automatic}. Much of the work uses a framework of six `universal' emotions \cite{ekman:1999}: sadness, happiness, fear, anger, surprise and disgust, with a further neutral category.

Deep learning models have been successful in automatically recognizing facial expressions in images~\cite{jung2015joint,liu2014facial,mollahosseini2016going,rodriguez2017deep,yu2015image,zhang2017facial,zhang2015learning}. They learn hierarchical structures from low- to high-level feature representations thanks to the complex, multi-layered architectures of neural networks. Kahou~\etal~\cite{kahou2013combining} applied convolutional neural networks (CNNs) to recognize facial expressions and won the 2013 Emotion Recognition in the Wild (EmotiW) Challenge. Another CNN model, followed by a linear support vector machine, was trained to recognize facial expressions by Tang~\etal~\cite{tang2013deep}; this won the 2013 Facial Expression Recognition (FER) challenge \cite{goodfellow2013challenges}. Kahou~\etal~\cite{kahou2016emonets} applied CNNs for extracting visual features accompanied by audio features in a multi-modal data representation. Nezami \etal~\cite{nezami2018face} used a CNN model to recognize facial expressions, where the learned representation is used in an image captioning model; the model embedded the recognized facial expressions to generate more human-like captions for images including human faces. Yu~\etal~\cite{yu2015image} employed a CNN model that was pre-trained on the FER-2013 dataset \cite{goodfellow2013challenges} and fine-tuned on the Static Facial Expression in the Wild (SFEW) dataset \cite{dhall2011static}. They applied a face detection method to detect faces and remove noise in their target data samples. Mollahosseini~\etal~\cite{mollahosseini2016going} trained CNN models across different well-known FER datasets to enhance the generalizablity of recognizing facial expressions. They applied face registration processes, extracting and aligning faces, to achieve better performances. Kim~\etal~\cite{kim2016fusing} measured the impact of combining registered and unregistered faces in this domain. They used the unregistered faces when the facial landmarks of the faces were not detectable. Zhang~\etal~\cite{zhang2017facial} applied CNNs to capture spatial information from video frames. The spatial information was combined with temporal information to recognize facial expressions. Pramerdorfer~\etal~\cite{pramerdorfer2016facial} employed a combination of modern deep architectures such as VGGnet \cite{simonyan2014very} on the FER-2013 dataset. They also achieved the state-of-the-art result on the FER-2013 dataset.

\subsection{Engagement Recognition}
Engagement has been detected in three different time scales: the entire video of a learning session, 10-second video clips and images. In the first category, Grafsgarrd~\etal~\cite{grafsgaard2013automatically} studied the relation between facial action units (AUs) and engagement in learning contexts. They collected videos of web-based sessions between students and tutors. After finishing the sessions, they requested each student to fill out an engagement survey used to annotate the student's engagement level. Then, they used linear regression methods to find the relationship between different levels of engagement and different AUs. However, their approach does not characterize engagement in fine-grained time intervals which are required for making an adaptive educational interface.

As an attempt to solve this issue, Whitehill~\etal~\cite{whitehill2014faces} applied linear support vector machines (SVMs) and Gabor features, as the best approach in this work, to classify four engagement levels: not engaged at all, nominally engaged, engaged in task, and very engaged. In this work, the dataset includes 10-second videos annotated into the four levels of engagement by observers, who are analyzing the videos. Monkaresi~\etal~\cite{monkaresi2017automated} used heart rate features in addition to facial features to detect engagement. They used a face tracking engine to extract facial features and WEKA (a classification toolbox) to classify the features into engaged or not engaged classes. They annotated their dataset, including 10-second videos, using self-reported data collected from students during and after their tasks. Bosch~\etal~\cite{bosch2015automatic} detected engagement using AUs and Bayesian classifiers. The generalizability of the model was also investigated across different times, days, ethnicities and genders \cite{bosch2016using}. Furthermore, in interacting with intelligent tutoring systems (ITSs), engagement was investigated based on a personalized model including appearance and context features \cite{alyuz2016semi}. Engagement was considered in learning with massively open online courses (MOOCs) as an e-learning environment \cite{d2016daisee}. In such settings, data are usually annotated by observing video clips or filling self-reports. However, the engagement levels of students can change during 10-second video clips, so assigning a single label to each clip is difficult and sometimes inaccurate.

In the third category, HOG features and SVMs have been applied to classify images using three levels of engagement: not engaged, nominally engaged and very engaged \cite{kamath2016crowdsourced}. This work is based on the experimental results of whitehill~\etal~\cite{whitehill2014faces} in preparing engagement samples. whitehill~\etal~\cite{whitehill2014faces} showed that engagement patterns are mostly recorded in images. Bosch~\etal~\cite{bosch2015automatic} also confirmed that video clips could not provide extra information by reporting similar performances using different lengths of video clips in detecting engagement.
However, competitive performances are not reported in this category.

We focus on the third category to recognize engagement from images. To do so, we collected a new dataset annotated by Psychology students, who can potentially better recognize the psychological phenomena of engagement, because of the complexity of analyzing student engagement. To assist them with recognition, brief training was provided prior to commencing the task and delivered in a consistent manner via online examples and descriptions. We did not use crowdsourced labels, resulting in less effective outcomes, similar to the work of Kamath~\etal~\cite{kamath2016crowdsourced}. Furthermore, we captured more effective labels by following an annotation process to simplify the engagement concept into the behavioral and the emotional dimensions. We requested annotators to label the dimensions for each image and make the overall annotation label by combining these. Our aim is for this dataset to be useful to other researchers interested in detecting engagement from images. Given this dataset, we introduce a novel model to recognize engagement using deep learning. Our model includes two important phases. First, we train a deep model to recognize basic facial expressions. Second, the model is applied to initialize the weights of our engagement recognition model trained using our newly collected dataset.

\begin{figure}[t]
    \begin{center}
    \includegraphics[width=0.6\linewidth]{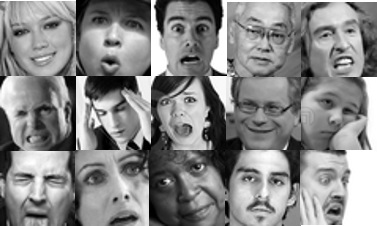}
    \end{center}
    \caption{Examples from the FER-2013 dataset~\cite{goodfellow2013challenges} including seven basic facial expressions.}
\label{fig:fer}
\end{figure}

\section{Facial Expression Recognition from Images}
\label{sec:facemodel}

\subsection{Facial Expression Recognition Dataset}
\label{sec:facedata}
In this section, we use the facial expression recognition 2013 (FER-2013) dataset~\cite{goodfellow2013challenges}. The dataset includes images, labeled \textit{happiness}, \textit{anger}, \textit{sadness}, \textit{surprise}, \textit{fear}, \textit{disgust}, and \textit{neutral}. It contains 35,887 samples (28,709 for the training set, 3589 for the public test set and 3589 for the private test set), collected by the Google search API. The samples are in grayscale at the size of 48-by-48 pixels (Fig.~\ref{fig:fer}).

We split the training set into two parts after removing 11 completely black samples: 3589 for validating and 25,109 for training our facial expression recognition model. To compare with related work~\cite{kim2016fusing,pramerdorfer2016facial,yu2015image}, we do not use the public test set for training or validation, but use the private test set for performance evaluation of our facial expression recognition model.

\begin{figure}[t]
    \begin{center}
    \includegraphics[width=0.34\linewidth]{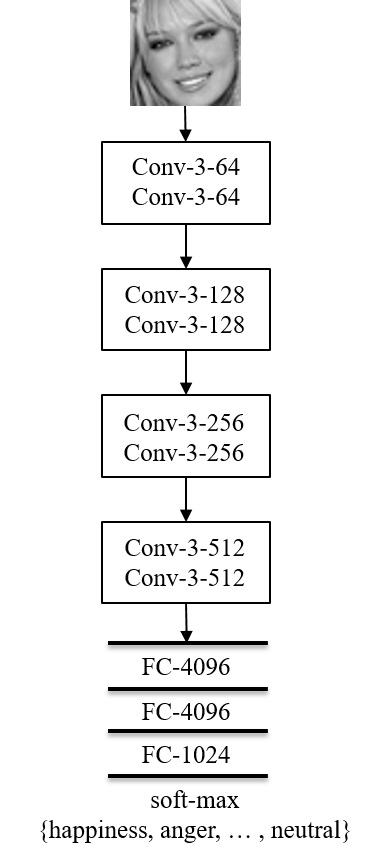}
    \end{center}
    \caption{The architecture of our facial expression recognition model adapted from VGG-B framework \cite{simonyan2014very}. Each rectangle is a Conv. block including two Conv. layers. The max pooling layers are not shown for simplicity.}
\label{fig:fermodel}
\end{figure}

\subsection{Facial Expression Recognition using Deep Learning}
\label{sec:pretrain}
We train the VGG-B model \cite{simonyan2014very}, using the FER-2013 dataset, with one less Convolutional (Conv.) block as shown in Fig.~\ref{fig:fermodel}. This results in eight Conv. and three fully connected layers. We also have a max pooling layer after each Conv. block with stride $2$. We normalize each FER-2013 image so that the image has a mean $0.0$ and a norm $100.0$ \cite{tang2013deep}. Moreover, for each pixel position, the pixel value is normalized to mean $0.0$ and standard-deviation $1.0$ using our training part. Our model has a similar performance with the work of Pramerdorfer~\etal~\cite{pramerdorfer2016facial} generating the state-of-the-art on FER-2013 dataset. The model's output layer has a softmax function generating the categorical distribution probabilities over seven facial expression classes in FER-2013. We aim to use this model as a part of our engagement recognition model.

\begin{figure}[t]
    \begin{center}
    \includegraphics[width=0.55\linewidth]{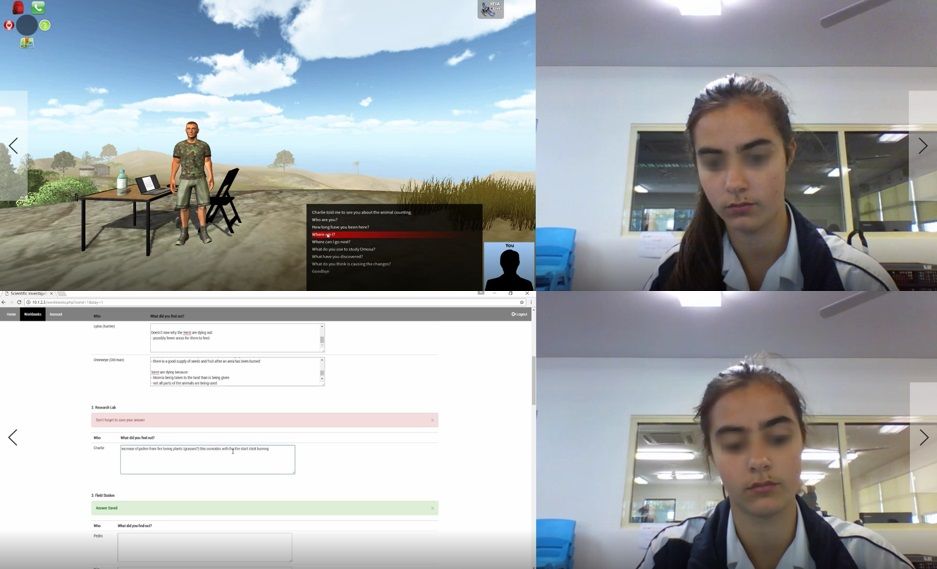}
    \end{center}
    \caption{The interactions of a student with Omosa \cite{jacobson2016computational}, captured in our studies.}
\label{fig:omos}
\end{figure}

\section{Engagement Recognition from Images}
\label{sec:engmodel}

\subsection{Engagement Recognition Dataset}
\label{sec:engdata}
\paragraph{Data Collection} To recognize engagement from face images, we construct a new dataset that we call the Engagement Recognition (ER) dataset. The data samples are extracted from videos of students, who are learning scientific knowledge and research skills using a virtual world named Omosa \cite{jacobson2016computational}. Samples are taken at a fixed rate instead of random selections, making our dataset samples representative, spread across both subjects and time. In the interaction with Omosa, the goal of students is to determine why a certain animal kind is dying out by talking to characters, observing the animals and collecting relevant information, Fig.~\ref{fig:omos} (top). After collecting notes and evidence, students are required to complete a workbook, Fig.~\ref{fig:omos} (bottom).

\begin{figure}[t]
    \begin{center}
    \includegraphics[width=1.0\linewidth]{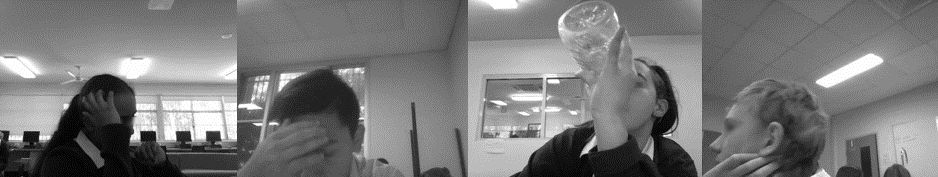}
    \end{center}
    \caption{Examples without detectable faces because of high face occlusions.}
\label{fig:un} 
\end{figure}

The videos of students were captured from our studies in two public secondary schools involving twenty students (11 girls and 9 boys) from Years 9 and 10 (aged 14--16), whose parents agreed to their participation in our ethics-approved studies. We collected the videos from twenty individual sessions of students recorded at 20 frames per second (fps), resulting in twenty videos and totalling around 20 hours. After extracting video samples, we applied a convolutional neural network (CNN) based face detection algorithm~\cite{king2009dlib} to select samples including detectable faces. The face detection algorithm cannot detect faces in a small number of samples (less than $1\%$) due to their high face occlusion (Fig.~\ref{fig:un}). We removed the occluded samples from the ER dataset.

\paragraph{Data Annotation} We designed custom annotation software to request annotators to independently label 100 samples each. The samples are randomly selected from our collected data and are displayed in different orders for different annotators. Each sample is annotated by at least six annotators.\footnote{The Fleiss' kappa of the six annotators is
0.59, indicating a high inter-coder agreement.} Following ethics approval, we recruited undergraduate Psychology students to undertake the annotation task, who received course credit for their participation. Before starting the annotation process, annotators were provided with definitions of behavioral and emotional dimensions of engagement, which are defined in the following paragraphs, inspired by the work of Aslan~\etal~\cite{aslan2017human}.

\vspace{0.25cm}

\noindent
\textit{Behavioral dimension}:
\begin{itemize}
    \item  \textit{On-Task}: The student is looking towards the screen or looking down to the keyboard below the screen.
    \item  \textit{Off-Task}: The student is looking everywhere else or eyes completely closed, or head turned away.
    \item  \textit{Can't Decide}: If you cannot decide on the behavioral state.
\end{itemize}

\begin{figure}[t]
    \begin{center}
    \includegraphics[width=0.9\linewidth]{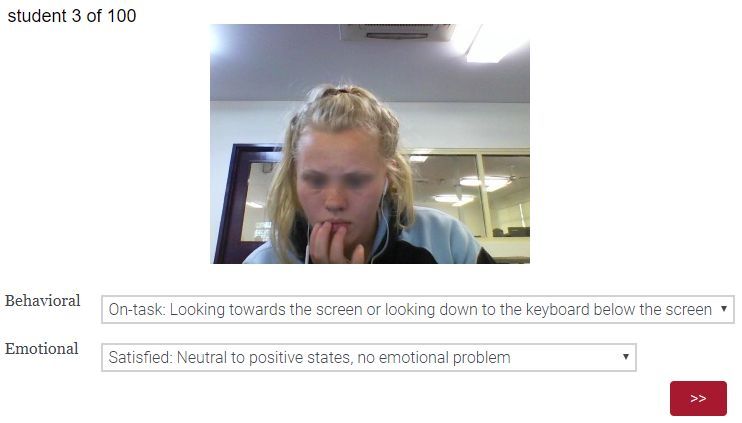}
    \end{center}
    \caption{An example of our annotation software where the annotator is requested to specify the behavioral and emotional dimensions of the displayed sample.}
\label{fig:software}
\end{figure}

\begin{table}
\caption{The adapted relationship between the behavioral and emotional dimensions from Woolf~\etal~\cite{woolf2009affect} and Aslan~\etal~\cite{aslan2017human}.}\label{tab:comb} 
\begin{center}
\begin{tabular}{|l|l|l|}
\hline
Behavioral & Emotional & Engagement  \\ \hline
On-task & Satisfied & Engaged \\
On-task & Confused & Engaged \\
On-task & Bored & Disengaged \\
Off-task & Satisfied & Disengaged \\
Off-task & Confused & Disengaged \\
Off-task & Bored & Disengaged \\
\hline
\end{tabular}
\end{center}
\end{table}

\noindent
\textit{Emotional dimension}:
\begin{itemize}
    \item  \textit{Satisfied}: If the student is not having any emotional problems during the learning task. This can include all positive states of the student from being neutral to being excited during the learning task.  
    \item  \textit{Confused}: If the student is getting confused during the learning task.
In some cases, this state might include some other negative states such as frustration.
    \item  \textit{Bored}: If the student is feeling bored during the learning task.
    \item  \textit{Can't Decide}: If you cannot decide on the emotional state.
\end{itemize}

During the annotation process, we show each data sample followed by two questions indicating the engagement's dimensions. The behavioral dimension can be chosen among \textit{on-task}, \textit{off-task}, and \textit{can't decide} options and the emotional dimension can be chosen among \textit{satisfied}, \textit{confused}, \textit{bored}, and \textit{can't decide} options. In each annotation phase, annotators have access to the definitions to label each dimension. A sample of the annotation software is shown in Fig.~\ref{fig:software}. In the next step, each sample is categorized as engaged or disengaged by combining the dimensions' labels using Table \ref{tab:comb}. For example, if a particular annotator labels an image as \textit{on-task} and \textit{satisfied}, the category for this image from this annotator is \textit{engaged}. Then, for each image we use the majority of the engaged and disengaged labels to specify the final overall annotation. If a sample receives the label of \textit{can't decide} more than twice (either for the emotional or behavioral dimensions) from different annotators, it is removed from ER dataset. Labeling this kind of samples is a difficult task for annotators, notwithstanding the good level of agreement that was achieved, and finding solutions to reduce the difficulty remains as a future direction of our work. Using this approach, we have created ER dataset consisting of 4627 annotated images including 2290 engaged and 2337 disengaged.

\begin{figure}[t]
    \begin{center}
    \includegraphics[width=0.90\linewidth]{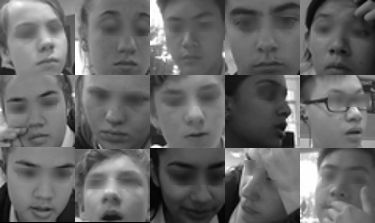}
    \end{center}
    \vspace{-0.3cm}
    \caption{Randomly selected images of ER dataset including engaged and disengaged.}
\label{fig:er}
\end{figure}

\paragraph{Dataset Preparation} We apply the CNN based face detection algorithm to detect the face of each ER sample. If there is more than one face in a sample, we choose the face with the biggest size. Then, the face is transformed to grayscale and resized into 48-by-48 pixels, which is an effective resolution for engagement detection \cite{whitehill2014faces}. Fig.~\ref{fig:er} shows some examples of the ER dataset. We split the ER dataset into training (3224), validation (715), and testing (688) sets, which are subject-independent (the samples in these three sets are from different subjects). Table \ref{tab:sat} demonstrates the statistics of these three sets.

\begin{table}
\caption{The statistics of ER dataset and its partitions.}
\vspace{-0.3cm}
\label{tab:sat}
\begin{center}
\begin{tabular}{|l|c|c|c|c|}
\hline
State & Total & Train & Valid & Test \\ \hline
Engaged & 2290 & 1589 & 392 & 309\\ 
Disengaged & 2337 & 1635 & 323 & 379\\
Total & 4627 & 3224 & 715 & 688 \\ \hline
\end{tabular}
\end{center}
 
\end{table}

\subsection{Engagement Recognition using Deep Learning}
\label{sec:EngDeep}

We define two Convolutional Neural Network (CNN) architectures as baselines, one designed architecture and one that is similar in structure to VGGnet \cite{simonyan2014very}. The key model of interest in this paper is a version of the latter baseline that incorporates facial expression recognition. For completeness, we also include another baseline that is not based on deep learning, but rather uses support vector machines (SVMs) with histogram of oriented gradients (HOG) features. For all the models, every sample of the ER dataset is normalized so that it has a zero mean and a norm equal to $100.0$. Furthermore, for each pixel location, the pixel values are normalized to mean zero and standard deviation one using all ER training data.

\paragraph{HOG+SVM} We trained a method using the histogram of oriented gradients (HOG) features extracted from ER samples and a linear support vector machine (SVM), which we call the \modelname{HOG+SVM model}. The model is similar to that of Kamath~\etal~\cite{kamath2016crowdsourced} for recognizing engagement from images and is used as a baseline model in this work. HOG \cite{dalal2005histograms} applies gradient directions or edge orientations to express objects in local regions of images. For example, in facial expression recognition tasks, HOG features can represent the forehead's wrinkling by horizontal edges. A linear SVM is usually used to classify HOG features. In our work, $C$, determining the misclassification rate of training samples against the objective function of SVM, is fine-tuned, using the validation set of the ER dataset, to the value of $0.1$.

\begin{figure}[t]
    \begin{center}
    \includegraphics[width=1.0\linewidth]{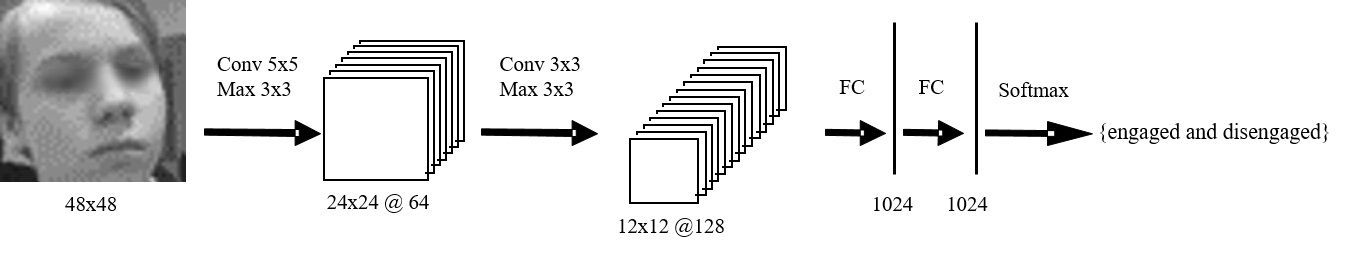}
    \end{center}
    \caption{The architecture of the \modelname{CNN Model}. We denote convolutional, max-pooling, and fully-connected layers with ``Conv'', ``Max'', and ``FC'', respectively.}
\label{fig:CNN}
\end{figure}

\paragraph{Convolutional Neural Network} We use the training and validation sets of the ER dataset to train a Convolutional Neural Networks (CNNs) for this task from scratch (the \modelname{CNN model}); this constitutes another of the baseline models in this paper. The model's architecture is shown in Fig.~\ref{fig:CNN}. The model contains two convolutional (Conv.) layers, followed by two max pooling (Max.) layers with stride 2, and two fully connected (FC) layers, respectively. A rectified linear unit (ReLU) activation function \cite{nair2010rectified} is applied after all Conv. and FC layers. The last step of the CNN model includes a softmax layer, followed by a cross-entropy loss, which consists of two neurons indicating engaged and disengaged classes. To overcome model over-fitting, we apply a dropout layer \cite{srivastava2014dropout} after every Conv. and hidden FC layer. Local response normalization \cite{krizhevsky2012imagenet} is used after the first Conv. layer. As the optimizer algorithm, stochastic gradient descent with mini-batching and a momentum of 0.9 is used. Using Equation \ref{eq:cnnlr}, the learning rate at step $t$ ($a_t$) is decayed by the rate ($r$) of 0.8 in the decay step ($s$) of 500. The total number of iterations from the beginning of the training phase is global step ($g$).

\begin{equation}
a_t=a_{t-1} \times r^ \frac{g}{s}
\label{eq:cnnlr}
\end{equation}

\begin{figure}[t]
\begin{minipage}[b]{0.47\linewidth}
\begin{center}
    \includegraphics[width=0.60\linewidth]{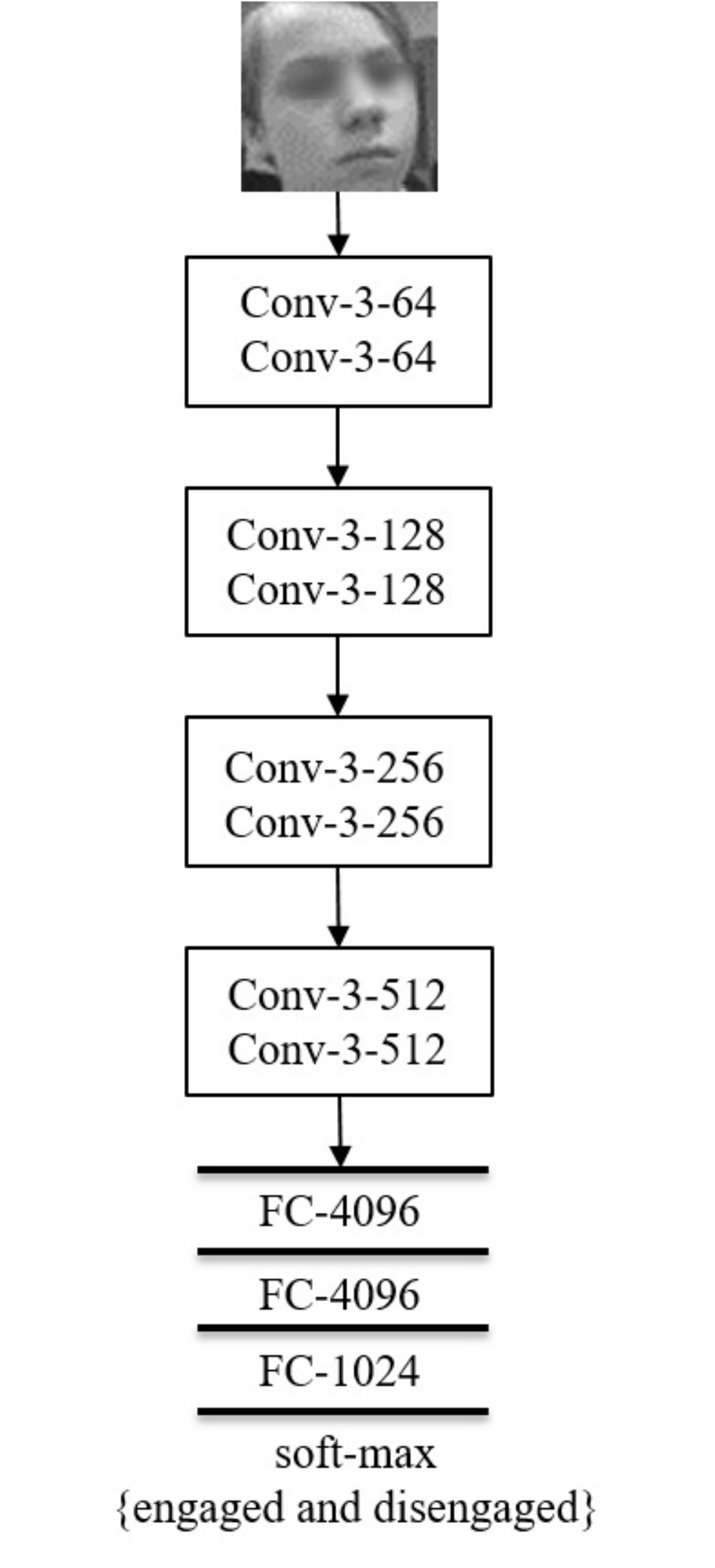}
    \end{center}
    \caption{The architecture of the \modelname{VGGnet model} on ER dataset. ``Conv'' and ``FC'' are convolutional and fully connected layers.}
\label{fig:vgg}
\end{minipage}
\hspace{0.06cm}
\begin{minipage}[b]{0.47\linewidth}
\begin{center}
    \includegraphics[width=1.20\linewidth]{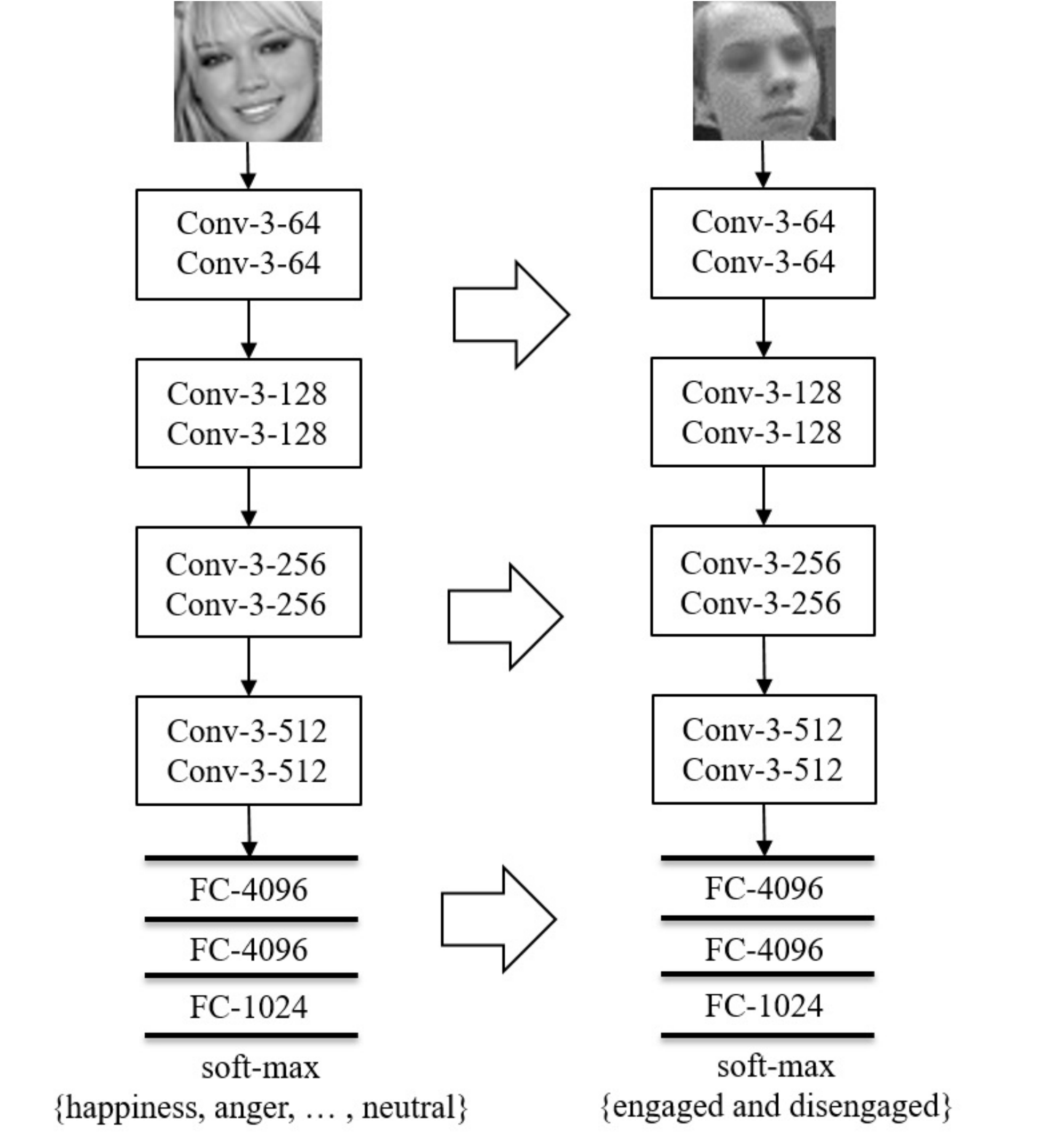}
    \end{center}
    \caption{Our facial expression recognition model on FER-2013 dataset (left). The \modelname{engagement model} on ER dataset (right).}
\label{fig:vggeng}
\end{minipage}
\end{figure}

\paragraph{Very Deep Convolutional Neural Network}
Using the ER dataset, we train a deep model which has eight Conv. and three FC layers similar to VGG-B architecture \cite{simonyan2014very}, but with two fewer Conv. layers. The model is trained using two different scenarios. Under the first scenario, the model is trained from scratch initialized with random weights; we call this the \modelname{VGGnet model} (Fig.~\ref{fig:vgg}), and this constitutes the second of our deep learning baseline models. Under the second scenario, which uses the same architecture, the model's layers, except the softmax layer, are initialized by the trained model of Section~\ref{sec:pretrain}, the goal of which is to recognize basic facial expressions; we call this the \modelname{engagement model} (Fig.~\ref{fig:vggeng}), and this is the key model of interest in our paper. In this model, all layers' weights are updated and fine-tuned to recognize engaged and disengaged classes in the ER dataset. For both \modelname{VGGnet} and \modelname{engagement models}, after each Conv. block, we have a max pooling layer with stride 2. In the models, the softmax layer has two output units (engaged and disengaged), followed by a cross-entropy loss. Similar to the \modelname{CNN model}, we apply a rectified linear unit (ReLU) activation function \cite{nair2010rectified} and a dropout layer \cite{srivastava2014dropout} after all Conv. and hidden FC layers. Furthermore, we apply local response normalization after the first Conv. block. We use the same approaches to optimization and learning rate decay as in the \modelname{CNN model}.

\section{Experiments}
\label{sec:exp}

\subsection{Evaluation Metrics}
\label{sec:EvalMetrics}
In this paper, the performance of all models are reported on the both validation and test splits of the ER dataset. We use three performance metrics including classification accuracy, F1 measure and the area under the ROC (receiver operating characteristics) curve (AUC). In this work, classification accuracy specifies the number of positive (engaged) and negative (disengaged) samples which are correctly classified and are divided by all testing samples (Equation \ref{eq:acc}).

\begin{equation}
Accuracy =\frac{TP+TN}{TP+FP+TN+FN}
\label{eq:acc}
\end{equation}

\noindent where $TP$, $TN$, $FP$, and $FN$ are true positive, true negative, false positive, and false negative, respectively. F1 measure is calculated using Equation \ref{eq:F1}.

\begin{equation}
F1 =2 \times \frac{p \times r}{p+r}
\label{eq:F1}
\end{equation}

\noindent where $p$ is precision defined as $\frac{TP}{TP+FP}$ and $r$ is recall defined as $\frac{TP}{TP+FN}$. AUC is a popular metric in engagement recognition task \cite{bosch2015automatic,monkaresi2017automated,whitehill2014faces}; it is an unbiased assessment of the area
under the ROC curve. An AUC score of 0.5 corresponds to chance performance by the classifier, and AUC 1.0 represents the best possible result.

\subsection{Implementation Details}
\label{sec:imdetail}
In the training phase, for data augmentation, input images are randomly flipped along their width and cropped to 48-by-48 pixels (after applying zero-padding because the samples were already in this size). Furthermore, they are randomly rotated by a specific max angle. We set learning rate for the \modelname{VGGnet model} to $0.001$ and for other models to $0.002$. The batch size is set to 32 for the \modelname{engagement model} and 28 for other models. The best model on the validation set is used to estimate the performance on the test partition of the ER dataset for all models in this work.

\subsection{Results}
\label{sec:res}
\paragraph{Overall Metrics} We summarize the experimental results on the validation set of the ER dataset in Table \ref{tab:resultsval} and on the test set of the ER dataset in Table \ref{tab:resultstest}. On the validation and test sets, the \modelname{engagement} model substantially outperforms all baseline models using all evaluation metrics, showing the effectiveness of using a trained model on basic facial expression data to initialize an engagement recognition model. All deep models including \modelname{CNN}, \modelname{VGGnet}, and \modelname{engagement models} perform better than the \modelname{HOG+SVM} method, showing the benefit of applying deep learning to recognize engagement. On the test set, the \modelname{engagement model} achieves $72.38\%$ classification accuracy, which outperforms \modelname{VGGnet} by 5\%, and the \modelname{CNN model} by more than $6\%$; it is also $12.5\%$ better than the \modelname{HOG+SVM} method. The \modelname{engagement model} achieved $73.90\%$ F1 measure which is around $3\%$ improvement compared to the deep baseline models and $6\%$ better performance than the \modelname{HOG+SVM model}. Using the AUC metric, as the most popular metric in engagement recognition tasks, the \modelname{engagement model} achieves $73.74\%$ which improves the \modelname{CNN} and \modelname{VGGnet models} by more than $5\%$ and is around $10\%$ better than the \modelname{HOG+SVM} method. There are similar improvements on the validation set.

\begin{table}[t]
\begin{minipage}[b]{0.45\linewidth}\centering
\caption{The results of our models (\%) on the validation set of ER dataset.}
\label{tab:resultsval}
\begin{center}
\begin{tabular}{|l|c|c|c|}
\hline
Method & Accuracy & F1 & AUC  \\ \hline
\modelname{HOG+SVM} & 67.69 & 75.40 & 65.50 \\
\modelname{CNN} & 72.03 & 74.94 & 71.56 \\
\modelname{VGGnet} & 68.11 & 70.69 & 67.85 \\
\modelname{engagement} & \textbf{77.76} & \textbf{81.18} & \textbf{76.77} \\
\hline
\end{tabular}
\end{center}

\end{minipage}
\hspace{0.5cm}
\begin{minipage}[b]{0.45\linewidth}
\caption{The results of our models (\%) on the test set of ER dataset.}
\label{tab:resultstest}
\begin{center}
\begin{tabular}{|l|c|c|c|}
\hline
Method & Accuracy & F1 & AUC  \\ \hline
\modelname{HOG+SVM} & 59.88 & 67.38 & 62.87 \\
\modelname{CNN} & 65.70 & 71.01 & 68.27 \\
\modelname{VGGnet} & 66.28 & 70.41 & 68.41 \\
\modelname{engagement} & \textbf{72.38} & \textbf{73.90} & \textbf{73.74} \\
\hline
\end{tabular}
\end{center}

\end{minipage}
\end{table}



\paragraph{Confusion Matrices} We show the confusion matrices of the \modelname{HOG+SVM}, \modelname{CNN}, \modelname{VGGnet}, and \modelname{engagement models} on the ER test set in Table~\ref{tab:confsvm}, Table~\ref{tab:confcnn}, Table~\ref{tab:confvgg}, and Table~\ref{tab:conftr}, respectively. The tables show the proportions of predicted classes with respect to the actual classes, allowing an examination of precision per class. It is interesting that the effectiveness of deep models comes through their ability to recognize disengaged samples compared to the \modelname{HOG+SVM model}. 

Disengaged samples have a wider variety of body postures and facial expressions than engaged sample 
(see engaged and disengaged examples in Fig.~11 in supplementary materials). 
Due to complex structures, deep learning models are more powerful in capturing these wider variations. The \modelname{VGGnet model}, which has a more complex architecture compared to the \modelname{CNN model}, can also detect disengaged samples with a higher probability. Since we pre-trained the \modelname{engagement model} on basic facial expression data including considerable variations of samples, this model is the most effective approach to recognize disengaged samples achieving $60.42\%$ precision which is around $27\%$ improvement in comparison with the \modelname{HOG+SVM model}
(See Fig.~12 and 13 in supplementary materials which are showing some challenging examples to recognize engagement).

\begin{table}[t]
\begin{minipage}[b]{0.45\linewidth}\centering
\caption{Confusion matrix of the \modelname{HOG+SVM model} (\%).}
\vspace{-0.8cm}
\label{tab:confsvm}
\begin{center}
\begin{tabular}{|l|l|c|c|}
\cline{3-4}
\multicolumn{1}{c}{} & & \multicolumn{2}{c|}{predicted} \\
\cline{3-4}
\multicolumn{1}{c}{} & & Engaged & Disengaged  \\
\hline
\multirow{2}{*}{actual} & Engaged & 92.23 & 7.77  \\
& Disengaged & 66.49 & 33.51  \\
\hline
\end{tabular}
\end{center}
\end{minipage}
\hspace{0.5cm}
\begin{minipage}[b]{0.45\linewidth}
\caption{Confusion matrix of the \modelname{CNN model} (\%).}
\vspace{-0.8cm}
\label{tab:confcnn}
\begin{center}
\begin{tabular}{|l|l|c|c|}
\cline{3-4}
\multicolumn{1}{c}{} & & \multicolumn{2}{c|}{predicted} \\
\cline{3-4}
\multicolumn{1}{c}{} & & Engaged & Disengaged  \\
\hline
\multirow{2}{*}{actual} & Engaged & 93.53 & 6.47  \\
& Disengaged & 56.99 & 43.01  \\
\hline
\end{tabular}
\end{center}
\end{minipage}
\end{table}

\begin{table}[t]
\begin{minipage}[b]{0.45\linewidth}\centering
\caption{Confusion matrix of the \modelname{VGGnet model} (\%).}
\vspace{-0.8cm}
\label{tab:confvgg}
\begin{center}
\begin{tabular}{|l|l|c|c|}
\cline{3-4}
\multicolumn{1}{c}{} & & \multicolumn{2}{c|}{predicted} \\
\cline{3-4}
\multicolumn{1}{c}{} & & Engaged & Disengaged  \\
\hline
\multirow{2}{*}{actual} & Engaged & 89.32 & 10.68  \\
& Disengaged & 52.51 & 47.49  \\
\hline
\end{tabular}
\end{center}
\end{minipage}
\hspace{0.5cm}
\begin{minipage}[b]{0.45\linewidth}
\caption{Confusion matrix of the \modelname{engagement model} (\%).}
\vspace{-0.8cm}
\label{tab:conftr}
\begin{center}
\begin{tabular}{|l|l|c|c|}
\cline{3-4}
\multicolumn{1}{c}{} & & \multicolumn{2}{c|}{predicted} \\
\cline{3-4}
\multicolumn{1}{c}{} & & Engaged & Disengaged  \\
\hline
\multirow{2}{*}{actual} & Engaged & 87.06 & 12.94  \\
& Disengaged & 39.58 & 60.42  \\
\hline
\end{tabular}
\end{center}
\end{minipage}
\end{table}

\section{Conclusion}
\label{sec:con}
Reliable models that can recognize engagement during a learning session, particularly in contexts where there is no instructor present, play a key role in allowing learning systems to intelligently adapt to facilitate the learner.  There is a shortage of data for training systems to do this; the first contribution of the paper is a new dataset, labelled by annotators with expertise in psychology, that we hope will facilitate research on engagement recognition from visual data. In this paper, we have used this dataset to train models for the task of automatic engagement recognition, including for the first time deep learning models. The next contribution has been the development of a model, called \modelname{engagement model}, that can address the shortage of engagement data to train a reliable deep learning model. \modelname{engagement model} has two key steps. First, we pre-train the model using basic facial expression data, of which is relatively abundant. Second, we train the model to produce a rich deep learning based representation for engagement, instead of commonly used features and classification methods in this domain. We have evaluated this model with respect to a comprehensive range of baseline models to demonstrate its effectiveness, and have shown that it leads to a considerable improvement against the baseline models using all standard evaluation metrics.

\pagebreak

\centering
{\large\bfseries SUPPLEMENTARY MATERIALS}

\setcounter{figure}{10}
\setcounter{page}{11}

\begin{figure*}
    \begin{center}
    \includegraphics[width=0.58\linewidth]{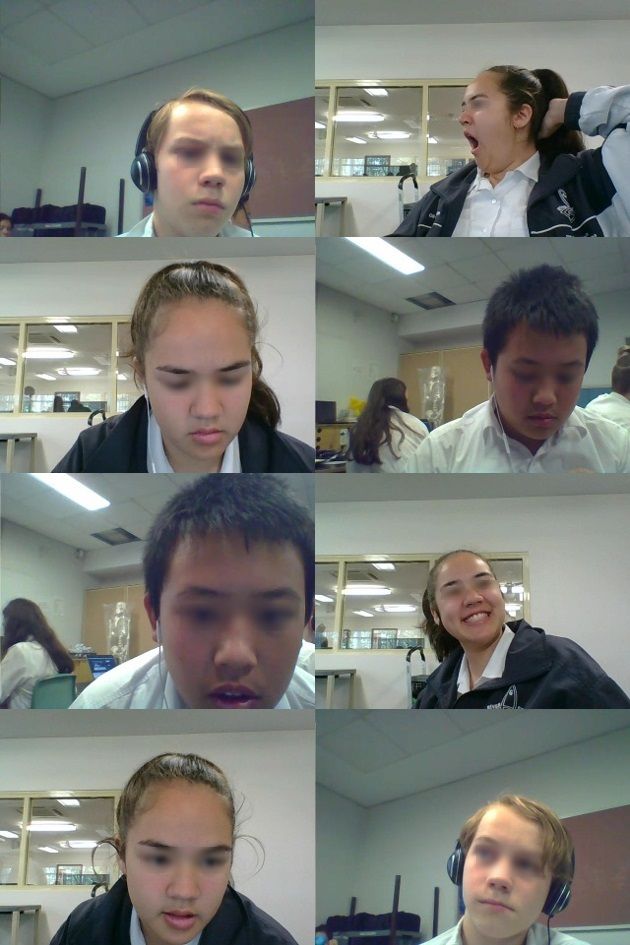}
    \end{center}
    \caption{Representative engaged (left) and disengaged (right) samples that are correctly classified using the \modelname{Transfer} model. Here, the label distribution across the engaged and disengaged classes is very different. This means that annotators can label these kinds of samples with less difficulties.}
\label{fig:prehigh}
\end{figure*}

\begin{figure*}
    \begin{center}
    \includegraphics[width=0.58\linewidth]{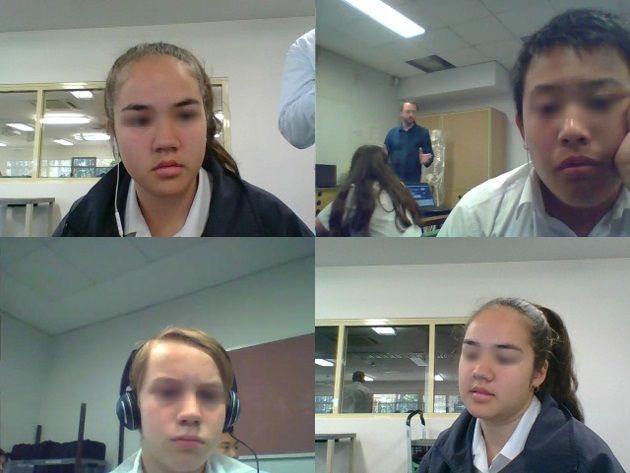}
    \end{center}
    \caption{Engaged (left) and disengaged (right) examples that are correctly detected using the \modelname{Transfer} model. Here, labeling these kinds of examples is difficult for annotators. As shown, there is more visual likeness between engaged and disengaged examples compared to the previous samples.}
\label{fig:preless}
\end{figure*}

\begin{figure*}
    \begin{center}
    \includegraphics[width=0.58\linewidth]{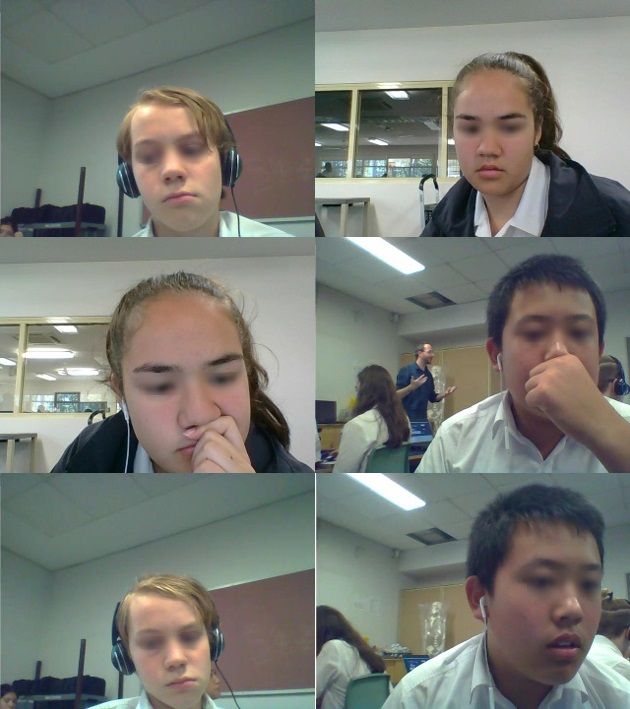}
    \end{center}
    \caption{These samples are wrongly predicted as engaged (left) and disengaged (right) using the \modelname{Transfer} model. The ground truth labels of the left samples are disengaged and the right samples are engaged. Here, labeling engaged or disengaged examples is also difficult for annotators which shows some challenging samples in engagement recognition tasks.}
\label{fig:prewrong}
\end{figure*}




%





\begin{thebibliography}{10}
\providecommand{\url}[1]{\texttt{#1}}
\providecommand{\urlprefix}{URL }
\providecommand{\doi}[1]{https://doi.org/#1}

\bibitem{alyuz2016semi}
Alyuz, N., Okur, E., Oktay, E., Genc, U., Aslan, S., Mete, S.E., Arnrich, B.,
  Esme, A.A.: Semi-supervised model personalization for improved detection of
  learner's emotional engagement. In: ICMI. pp. 100--107. ACM (2016)

\bibitem{aslan2017human}
Aslan, S., Mete, S.E., Okur, E., Oktay, E., Alyuz, N., Genc, U.E., Stanhill,
  D., Esme, A.A.: Human expert labeling process (help): Towards a reliable
  higher-order user state labeling process and tool to assess student
  engagement. Educational Technology pp. 53--59 (2017)

\bibitem{bosch2016detecting}
Bosch, N.: Detecting student engagement: Human versus machine. In: UMAP. pp.
  317--320. ACM (2016)

\bibitem{bosch2015automatic}
Bosch, N., D'Mello, S., Baker, R., Ocumpaugh, J., Shute, V., Ventura, M., Wang,
  L., Zhao, W.: Automatic detection of learning-centered affective states in
  the wild. In: IUI. pp. 379--388. ACM (2015)

\bibitem{bosch2016using}
Bosch, N., D'mello, S.K., Ocumpaugh, J., Baker, R.S., Shute, V.: Using video to
  automatically detect learner affect in computer-enabled classrooms. ACM
  Transactions on Interactive Intelligent Systems  \textbf{6}(2), ~17 (2016)

\bibitem{dalal2005histograms}
Dalal, N., Triggs, B.: Histograms of oriented gradients for human detection.
  In: CVPR. vol.~1, pp. 886--893. IEEE (2005)

\bibitem{d2016daisee}
D'Cunha, A., Gupta, A., Awasthi, K., Balasubramanian, V.: Daisee: Towards user
  engagement recognition in the wild. arXiv preprint arXiv:1609.01885  (2016)

\bibitem{dhall2011static}
Dhall, A., Goecke, R., Lucey, S., Gedeon, T.: Static facial expression analysis
  in tough conditions: Data, evaluation protocol and benchmark. In: ICCV. pp.
  2106--2112 (2011)

\bibitem{ekman:1999}
Ekman, P.: {Basic emotions}. In: Dalgleish, T., Power, T. (eds.) The Handbook
  of Cognition and Emotion, pp. 45--60. John Wiley \& Sons, Sussex, UK (1999)

\bibitem{ekman2006darwin}
Ekman, P.: Darwin and facial expression: A century of research in review. Ishk
  (2006)

\bibitem{fasel2003automatic}
Fasel, B., Luettin, J.: Automatic facial expression analysis: a survey. Pattern
  recognition  \textbf{36}(1),  259--275 (2003)

\bibitem{goodfellow2013challenges}
Goodfellow, I.J., Erhan, D., Carrier, P.L., Courville, A., Mirza, M., Hamner,
  B., Cukierski, W., Tang, Y., Thaler, D., Lee, D.H., et~al.: Challenges in
  representation learning: A report on three machine learning contests. In:
  ICONIP. pp. 117--124. Springer (2013)

\bibitem{grafsgaard2013automatically}
Grafsgaard, J., Wiggins, J.B., Boyer, K.E., Wiebe, E.N., Lester, J.:
  Automatically recognizing facial expression: Predicting engagement and
  frustration. In: Educational Data Mining 2013 (2013)

\bibitem{jacobson2016computational}
Jacobson, M.J., Taylor, C.E., Richards, D.: Computational scientific inquiry
  with virtual worlds and agent-based models: new ways of doing science to
  learn science. Interactive Learning Environments  \textbf{24}(8),  2080--2108
  (2016)

\bibitem{jung2015joint}
Jung, H., Lee, S., Yim, J., Park, S., Kim, J.: Joint fine-tuning in deep neural
  networks for facial expression recognition. In: ICCV. pp. 2983--2991 (2015)

\bibitem{kahou2016emonets}
Kahou, S.E., Bouthillier, X., Lamblin, P., Gulcehre, C., Michalski, V., Konda,
  K., Jean, S., Froumenty, P., Dauphin, Y., Boulanger-Lewandowski, N., et~al.:
  Emonets: Multimodal deep learning approaches for emotion recognition in
  video. Journal on Multimodal User Interfaces  \textbf{10}(2),  99--111 (2016)

\bibitem{kahou2013combining}
Kahou, S.E., Pal, C., Bouthillier, X., Froumenty, P., G{\"u}l{\c{c}}ehre,
  {\c{C}}., Memisevic, R., Vincent, P., Courville, A., Bengio, Y., Ferrari,
  R.C., et~al.: Combining modality specific deep neural networks for emotion
  recognition in video. In: ICMI. pp. 543--550. ACM (2013)

\bibitem{kamath2016crowdsourced}
Kamath, A., Biswas, A., Balasubramanian, V.: A crowdsourced approach to student
  engagement recognition in e-learning environments. In: WACV. pp.~1--9. IEEE
  (2016)

\bibitem{kapoor2001towards}
Kapoor, A., Mota, S., Picard, R.W., et~al.: Towards a learning companion that
  recognizes affect. In: AAAI Fall symposium. pp.~2--4 (2001)

\bibitem{kim2016fusing}
Kim, B.K., Dong, S.Y., Roh, J., Kim, G., Lee, S.Y.: Fusing aligned and
  non-aligned face information for automatic affect recognition in the wild: A
  deep learning approach. In: CVPR Workshops. pp. 48--57. IEEE (2016)

\bibitem{king2009dlib}
King, D.E.: Dlib-ml: A machine learning toolkit. Journal of Machine Learning
  Research  \textbf{10}(Jul),  1755--1758 (2009)

\bibitem{krizhevsky2012imagenet}
Krizhevsky, A., Sutskever, I., Hinton, G.E.: Imagenet classification with deep
  convolutional neural networks. In: NIPS. pp. 1097--1105 (2012)

\bibitem{liu2014facial}
Liu, P., Han, S., Meng, Z., Tong, Y.: Facial expression recognition via a
  boosted deep belief network. In: CVPR. pp. 1805--1812 (2014)

\bibitem{mollahosseini2016going}
Mollahosseini, A., Chan, D., Mahoor, M.H.: Going deeper in facial expression
  recognition using deep neural networks. In: WACV. pp. 1--10. IEEE (2016)

\bibitem{monkaresi2017automated}
Monkaresi, H., Bosch, N., Calvo, R.A., D'Mello, S.K.: Automated detection of
  engagement using video-based estimation of facial expressions and heart rate.
  IEEE Transactions on Affective Computing  \textbf{8}(1),  15--28 (2017)

\bibitem{nair2010rectified}
Nair, V., Hinton, G.E.: Rectified linear units improve restricted boltzmann
  machines. In: ICML. pp. 807--814 (2010)

\bibitem{nezami2018face}
Nezami, O.M., Dras, M., Anderson, P., Hamey, L.: Face-cap: Image captioning
  using facial expression analysis. In: Joint European Conference on Machine
  Learning and Knowledge Discovery in Databases. pp. 226--240. Springer (2018)

\bibitem{o2016theoretical}
O'Brien, H.: Theoretical perspectives on user engagement. In: Why Engagement
  Matters, pp. 1--26. Springer (2016)

\bibitem{pramerdorfer2016facial}
Pramerdorfer, C., Kampel, M.: Facial expression recognition using convolutional
  neural networks: State of the art. arXiv preprint arXiv:1612.02903  (2016)

\bibitem{rodriguez2017deep}
Rodriguez, P., Cucurull, G., Gonzalez, J., Gonfaus, J.M., Nasrollahi, K.,
  Moeslund, T.B., Roca, F.X.: Deep pain: Exploiting long short-term memory
  networks for facial expression classification. IEEE Transactions on
  Cybernetics (99),  1--11 (2017)

\bibitem{sariyanidi2015automatic}
Sariyanidi, E., Gunes, H., Cavallaro, A.: Automatic analysis of facial affect:
  A survey of registration, representation, and recognition. IEEE Transactions
  on Pattern Analysis and Machine Intelligence  \textbf{37}(6),  1113--1133
  (2015)

\bibitem{simonyan2014very}
Simonyan, K., Zisserman, A.: Very deep convolutional networks for large-scale
  image recognition. arXiv preprint arXiv:1409.1556  (2014)

\bibitem{srivastava2014dropout}
Srivastava, N., Hinton, G., Krizhevsky, A., Sutskever, I., Salakhutdinov, R.:
  Dropout: A simple way to prevent neural networks from overfitting. The
  Journal of Machine Learning Research  \textbf{15}(1),  1929--1958 (2014)

\bibitem{tang2013deep}
Tang, Y.: Deep learning using linear support vector machines. arXiv preprint
  arXiv:1306.0239  (2013)

\bibitem{whitehill2014faces}
Whitehill, J., Serpell, Z., Lin, Y.C., Foster, A., Movellan, J.R.: The faces of
  engagement: Automatic recognition of student engagement from facial
  expressions. IEEE Transactions on Affective Computing  \textbf{5}(1),  86--98
  (2014)

\bibitem{woolf2009affect}
Woolf, B., Burleson, W., Arroyo, I., Dragon, T., Cooper, D., Picard, R.:
  Affect-aware tutors: recognising and responding to student affect.
  International Journal of Learning Technology  \textbf{4}(3-4),  129--164
  (2009)

\bibitem{yu2015image}
Yu, Z., Zhang, C.: Image based static facial expression recognition with
  multiple deep network learning. In: ICMI. pp. 435--442. ACM (2015)

\bibitem{zhang2017facial}
Zhang, K., Huang, Y., Du, Y., Wang, L.: Facial expression recognition based on
  deep evolutional spatial-temporal networks. IEEE Transactions on Image
  Processing  \textbf{26}(9),  4193--4203 (2017)

\bibitem{zhang2015learning}
Zhang, Z., Luo, P., Loy, C.C., Tang, X.: Learning social relation traits from
  face images. In: ICCV. pp. 3631--3639 (2015)

\end{thebibliography}
\end{document}